\documentclass[sigconf]{acmart}
\usepackage{graphicx} 
\usepackage{float}
\usepackage{booktabs} 
\usepackage{enumitem,kantlipsum}
\usepackage{threeparttable}
\usepackage{algorithm}
\usepackage{algpseudocode}
\usepackage{multirow}
\usepackage{array}
\usepackage{booktabs}
\usepackage{subfig}
\usepackage{arydshln}
\usepackage{pgfplots}
\usepackage{tikz}

\AtBeginDocument{%
  \providecommand\BibTeX{{%
    \normalfont B\kern-0.5em{\scshape i\kern-0.25em b}\kern-0.8em\TeX}}}

\acmSubmissionID{648}

\begin{document}

\title{Supervised Advantage Actor-Critic for \\ Recommender Systems}

\affiliation{%
  \institution{\vspace{0.5cm}}
}
 \author{Xin Xin}
 \affiliation{%
   \institution{School of Computer Science, Shandong University, China}
 }
 \email{xinxin@sdu.edu.cn}

 \author{Alexandros Karatzoglou}
 \affiliation{%
   \institution{Google Research, London, UK}
 }
 \email{alexkz@google.com}

 \author{Ioannis Arapakis}
 \affiliation{%
   \institution{Telefonica Research, Barcelona, Spain}
 }
 \email{ioannis.arapakis@telefonica.com}

 \author{Joemon M. Jose}
 \affiliation{%
   \institution{School of Computing Science, University of Glasgow, UK}
 }
 \email{Joemon.Jose@glasgow.ac.uk}

\begin{abstract}
Casting session-based or sequential recommendation as reinforcement learning (RL) through reward signals is a promising research direction towards recommender systems (RS) that maximize cumulative profits. However, the direct use of RL algorithms in the RS setting is impractical due to challenges like off-policy training, huge action spaces and lack of sufficient reward signals. Recent RL approaches for RS attempt to tackle these challenges by combining RL and (self-)supervised sequential learning, but still suffer from certain limitations. For example, the estimation of Q-values tends to be biased toward positive values due to the lack of negative reward signals. Moreover, the Q-values also depend heavily on the specific timestamp of a sequence.

To address the above problems, we propose negative sampling strategy for training the RL component and combine it with supervised sequential learning. We call this method Supervised Negative Q-learning (SNQN). Based on sampled (negative) actions (items), we can calculate the ``advantage'' of a positive action over the average case, which can be further utilized as a normalized weight for learning the supervised sequential part. This leads to another learning framework: Supervised Advantage Actor-Critic (SA2C). We instantiate SNQN and SA2C with four state-of-the-art sequential recommendation models and conduct experiments on two real-world datasets. Experimental results show that the proposed approaches achieve significantly better performance than state-of-the-art supervised methods and existing self-supervised RL methods . Code will be open-sourced.
\end{abstract}



\maketitle

\section{Introduction}
Over the last 20 years, users have been navigating online services such as, e-commerce \cite{reinforce-e-commerce}, video platforms, and music apps \cite{nextitnet} with the help of RS. Most of these use cases involve session-based/next-item  recommendation, in which recommendation are generated from the sequence of user interactions.

Session-based recommendation models can be trained in a (self-)supervised learning fashion, in which a sequential model (e.g., a transformer \cite{SASRec, Transformer} or a RNN \cite{gru4rec}) is trained to predict the next item in the sequence itself, rather than some ``external'' labels \cite{gru4rec,nextitnet,SASRec}. This training approach is also widely adopted in language modeling tasks, to predict the next word given the previous word sequence \cite{word2vec}. Supervised learning can lead to sub-optimal recommendations, since the loss function used in supervised learning is purely defined on the discrepancy between model predictions and the actual interactions in the sequence. Recommmendations from a model trained on such a loss function may not match the desired properties of a RS from the perspective of both users and service providers. For example, service providers may want to promote recommendations that can lead to real purchases not just clicks. Other desirable properties  like diversity and novelty of the recommended item lists, could be considered,
which leads to a multi-objective optimization problem
\cite{ribeiro2014multiobjective,paroto-efficient-zhang}. Recommendation models trained with simple supervised learning do not tackle the above expectations and objectives.

Reinforcement learning (RL) has achieved success in game control \cite{alphogo,humanlevelcontrol,srinivas2020curl,bcq}, robotics \cite{kober2013reinforcement-robot-survey} and related fields. Unlike game control and robotics, directly utilizing RL for RS comes with sets of unique difficulties and challenges. Model-free RL algorithms train the agent through an “error-and-correction'' manner, in which the RL agent needs to interact with the environment and collect experience. The training procedure forces the agent to imitate good actions and avoid bad ones. Applying this in RS is  problematic, since interactions with an under-trained policy would negatively affect the user experience. A typical solution is to perform off-policy learning from the logged implicit feedback data \cite{googlewsdmoffpolicycorrection,xin2020self}. This entails trying to infer a target policy from the data generated by a different behavior policy, which is still an open research problem due to its high variance \cite{munos2016safeandefficient}. Moreover, learning from implicit feedback also introduces the challenge of insufficient negative signal \cite{xin2020self,rendle2009bpr}. Another alternative is to use model-based RL algorithms, in which
a model is firstly constructed to simulate the environment (users). Then the agent can learn from the interactions with the simulated environment \cite{chen2019generativeusermodel,shi2019virtual}. However, these two-stage methods depend heavily on the accuracy of the constructed simulator.

Self-supervised reinforcement learning  \cite{xin2020self} has been proposed for RS, achieving promising results on off-line evaluation metrics. Two learning frameworks namely Self-Supervised Q-learning (SQN) and Self-Supervised Actor-Critic (SAC) are proposed. The key insight of self-supervised RL is to utilize the RL component as a form of a regularizer to fine-tune the recommendation model towards the defined rewards, for instance in the e-commerce domain provide recommendations that lead to more purchases rather than just clicks \cite{xin2020self}. Although SQN and SAC achieve good performance, they still suffer from some limitations. For example, the RL head\footnote{For simplicity, we make ``head'' and ``output layer'' interchangeable in this paper.} in SQN and SAC is only defined on positive (interacted) actions (items), so the negative comparison signals only come from the cross-entropy loss of the supervised part. As a result, the RL head contributes to reward-based learning but cannot be used to generate recommendations, as it lacks negative feedback to remove the bias introduced by the existence of only positive reward signals. 
SAC uses the output Q-values\footnote{The Q-value for a state and action is an estimate of the expected cumulative reward under this state-action pair.} as the critic to re-weight the actor (supervised part). 
Q-values depend heavily on the specific timestamp of a sequence, which introduces further bias to the learning procedure.

To address the above issues, we first propose a negative sampling strategy for training RL in a RS setting and then combine it with supervised sequential learning. We call this Supervised Negative Q-learning (SNQN). Another interpretation of negative sampling in RL is imitation learning under sparse reward settings \cite{reddy2019sqil}. Different from SQN, which only performs RL on positive actions (clicks, views, etc.), the RL output head of SNQN is learned on both positive actions and a set of sampled negative actions. This design allows the RL part of the SNQN to not only act as a regularizer but also as a good ranking model, that can also be used to generate recommendations. Based on the sampled negative actions and the estimate of the Q-values, we can calculate the ``advantage'' of a positive action over the other actions. We propose the Supervised Advantage Actor-Critic (SA2C), that uses this advantage instead of the raw Q-values to re-weight the supervised output layer. The advantage values can be seen as normalized Q-values that help us alleviate the bias from sequence timestamp on the estimation of Q-values. This work makes the following contributions:

    {\it 1)} We propose SNQN introducing negative sampling for the RL training of the RS model and then combine it with supervised sequential learning. Both the supervised head and the RL head can be used to generate recommendations. We show that joint training of the two heads with a shared base model helps to achieve better performance than separate learning.
     {\it 2)} We propose SA2C to calculate the {\it advantage} of a positive action. This advantage can be seen as a normalized Q-value and is further utilized to re-weight the supervised component.
    {\it 3)} We integrate the proposed SNQN and SA2C with four state-of-the-art recommendation models and conduct experiments on two real-world e-commerce datasets. Experimental results demonstrate the proposed methods are effective in improving the performance of RS compared to existing methods.

\section{Related Work}

 \label{related-work}
Recurrent neural networks (RNN) and convolutional networks (CNN) have shown promising results in modeling recommendation sequences \cite{gru4rec,caser-rec,nextitnet}. Transformer architectures have been proven to be highly successful  \cite{Transformer} for language modeling tasks, and self-attention for recommendations has received a lot of attention \cite{SASRec}.

RL has been previously applied in RS. \citet{googlewsdmoffpolicycorrection} proposed to calculate a propensity score to perform off-policy correction for off-policy learning. Model-based RL approaches \cite{chen2019generativeusermodel,didienvironment,jdkdd19}
attempt to eliminate the off-policy issue by building a model to simulate the environment. The policy can then be trained through interactions with the simulator. Two-stage approaches depend heavily on the accuracy of the simulator. Although related methods, such as generative adversarial networks (GANs) \cite{GAN}, achieve good performance when generating content like images and speeches, simulating users' responses is a much more complex and difficult task \cite{chen2019generativeusermodel}. 

Recently, \citet{xin2020self} proposed self-supervised reinforcement learning for RS. Two learning frameworks SQN and SAC are suggested. SQN augments the recommendation model with two heads. One is defined on the supervised mode and the other 
RL head is based on the Q-learning for positive reward actions. SQN co-trains the supervised loss and RL loss to conduct transfer learning between each other \cite{xin2020self}. Long term rewards e.g. a purchase at the end of a session can be incorporated into the learning process, while the model is still trained efficiently on logged data. As the Q-values are an estimation of the goodness of the actions, SAC further utilizes these Q-values to re-weight the supervised part. SQN and SAC can be seen as  attempts to  utilize a Q-learning based RL estimator to ``reinforce'' session-based supervised recommendation models \cite{xin2020self} and 
achieve promising results on off-line evaluation metrics.

Research on slate-based recommendation has also been conducted in \cite{ie2019slateq,gong2019exactk,googlewsdmoffpolicycorrection,chen2019generativeusermodel}, where actions are considered to be sets (slates) of items. This setting leads to an exponentially increased action space. Finally, bandit algorithms are also reward-driven and have long-term optimization perspective. However, bandit algorithms assume that taking actions does not affect the state \cite{li2010contextualbandit}, while actually recommendations do have an effect on user behavior \cite{rohde2018recogym}; hence RL is a more suitable choice for the RS task. Another related field is imitation learning, where the policy is learned from expert demonstrations \cite{GAIL,ho2016model-free-imitation,torabi2018behavioral-clone, reddy2019sqil}. 
\section{Method}
Let $\mathcal{I}$ denote the item set, then a user-item interaction sequence\footnote{In the real-world scenario, there may be different kinds of interactions. For instance, in e-commerce, the interactions can be clicks, purchases, add to basket and so on. In video platforms, the interactions can be characterized by the watching time of a video.} can be represented as $x_{1:t}=\left\{x_1,x_2,...,x_{t-1},x_t\right\}$, where $x_i \in \mathcal{I} (0< i \leq t)$ denotes the interacted item at timestamp $i$.  The task of next-item recommendation is to recommend the most relevant item $x_{t+1}$ to the user, given the sequence of $x_{1:t}$. 

A common solution is to build a recommendation model whose output is the classification logits $\mathbf{y}_{t+1}=[y_1,y_2,...y_n] \in \mathbb{R}^n$, where $n$ is the number of candidate items. Each candidate item corresponds to a class. The recommendation list for timestamp $t+1$ can be generated by choosing top-$k$ items according to $\mathbf{y}_{t+1}$. Typically one can use a generative sequential model $G(\cdot)$ to encode the input sequence into a hidden state $\mathbf{s}_t$ as $\mathbf{s}_t=G(x_{1:t})$. Generally speaking, plenty of deep-learning based models \cite{caser-rec,gru4rec,nextitnet,SASRec} can serve as the generative model $G(\cdot)$.
After that, a decoder can be utilized to map the hidden state to the classification logits as $\mathbf{y}_{t+1}=f(\mathbf{s}_t)$. It is usually defined as a simple fully connected layer or the inner product with candidate item embeddings \cite{nextitnet,gru4rec,caser-rec,SASRec}. 

\subsection{Reinforcement Learning Setup} 
\label{RL}
From the perspective of RL, the next item recommendation task can be formulated as a Markov Decision Process (MDP) \cite{shani2005mdp}, in which the recommendation agent interacts with the environments $\mathcal{E}$ (users) by sequentially recommending items to maximize the discounted cumulative rewards. The MDP can be defined by tuples 
of $(\mathcal{S},\mathcal{A},\mathbf{P},R, \rho_0,\gamma)$ \cite{xin2020self,chen2019generativeusermodel,googlewsdmoffpolicycorrection} where
\begin{itemize}
    \item $\mathcal{S}$: a continuous state space to describe the user state. The state of a user at timestamp $t$ can be represented as $\mathbf{s}_t=G(x_{1:t}) \in \mathcal{S}$ $(t>0)$.
    \item $\mathcal{A}$: a discrete action space which contains candidate items. The action $a$ of the agent is to recommend the selected item. In off-line training data, we can get the positive action at timestamp $t$ from the input sequence (i.e., $a_t^+=x_{t+1} (t\geq0)$).
    \item $\mathbf{P}$: $\mathcal{S} \times \mathcal{A} \times \mathcal{S} \rightarrow \mathbb{R}$ is the state transition probability. When learning from off-line data, we can make an assumption that only positive actions can affect the user state. In other words, taking a negative (unobserved) action doesn't update the user state \cite{pairwise-q-learning,ie2019slateq}.
    \item $R$: $\mathcal{S} \times \mathcal{A} \rightarrow \mathbb{R}$ is the reward function, where $r(\mathbf{s},a)$ denotes the immediate reward by taking action $a$ at state $\mathbf{s}$. The flexible reward scheme allows the agent to optimize the recommendation models towards expectations that are not captured by simple supervised loss functions.
    \item $\rho_0$ is the initial state distribution with $\mathbf{s}_0 \sim \rho_0$.
    \item $\gamma$ is the discount factor for future rewards.
\end{itemize}
The goal of RL is to seek a target policy $\pi_\theta(a|\mathbf{s})$ so that sampling trajectories according to $\pi_\theta(a|\mathbf{s})$, would lead to the maximum expected cumulative reward:
\begin{equation}
	\label{cumulative-rewards}
	\max_{\pi_\theta}\mathbb{E}_{\tau\sim\pi_\theta}[R(\tau)]\text{, where }R(\tau)=\sum_{t=0}^{|\tau|}\gamma^{t}r(\mathbf{s}_t,a_t),
\end{equation}
where $\theta \in \mathbb{R}^d$ denotes policy parameters. Note that the expectation is taken over trajectories $\tau=(\mathbf{s}_0,a_0,\mathbf{s}_1,...)$, which are obtained by performing actions according to the target policy.

In on-line RL environments like game control, it's easy to sample the trajectories $\tau\sim\pi_\theta$ and the agent is trained through an “error-and-correction'' approach. However, under the RS setting, we cannot afford to make “errors'' (i.e. letting the user interact with under-trained policies) due to the negative impact on the user experience. Even if we can split a small portion of traffic to make the RL agent interact with live users, the final recommended items may still be controlled by other recommenders with different policies, since many recommendation models are deployed in a real-live RS. As a result, the sampled  trajectories will come from another behavior policy $\tau\sim\beta$ and we will resort to off-policy RL \cite{munos2016safeandefficient,googlewsdmoffpolicycorrection} and in particular Q-learning \cite{alphogo}.

\begin{figure*}
    \captionsetup[subfloat]
    {}
    \centering
    \subfloat[SNQN architecture.]{%
    \label{fig:snqn-architecture}
    \includegraphics[width=0.33\textwidth]{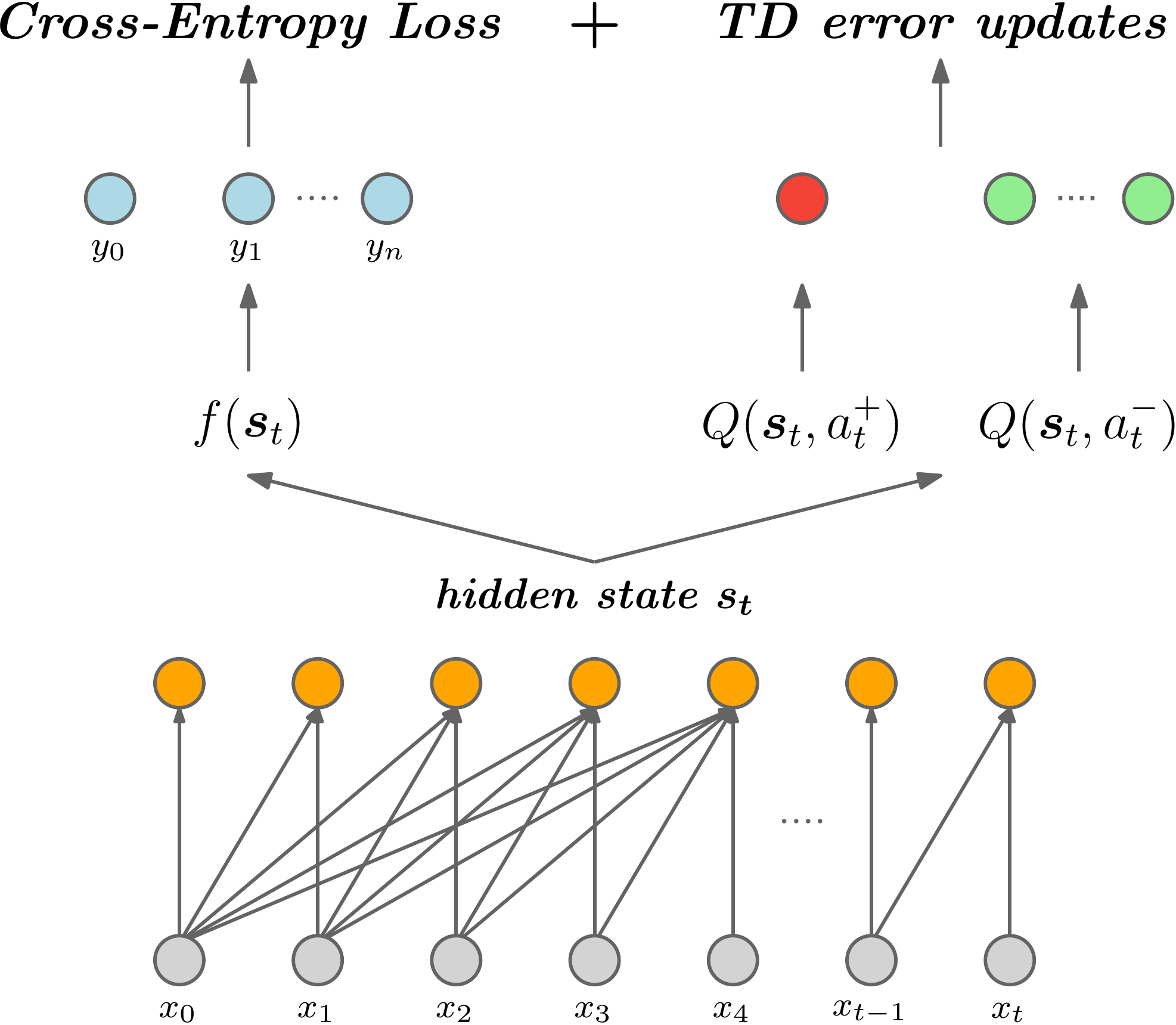}}
    \hspace{2cm}
    \subfloat[SA2C architecture. CE is short for cross-entropy. ]{
    \label{fig:sa2c-architecture}
    \includegraphics[width=0.33\textwidth]{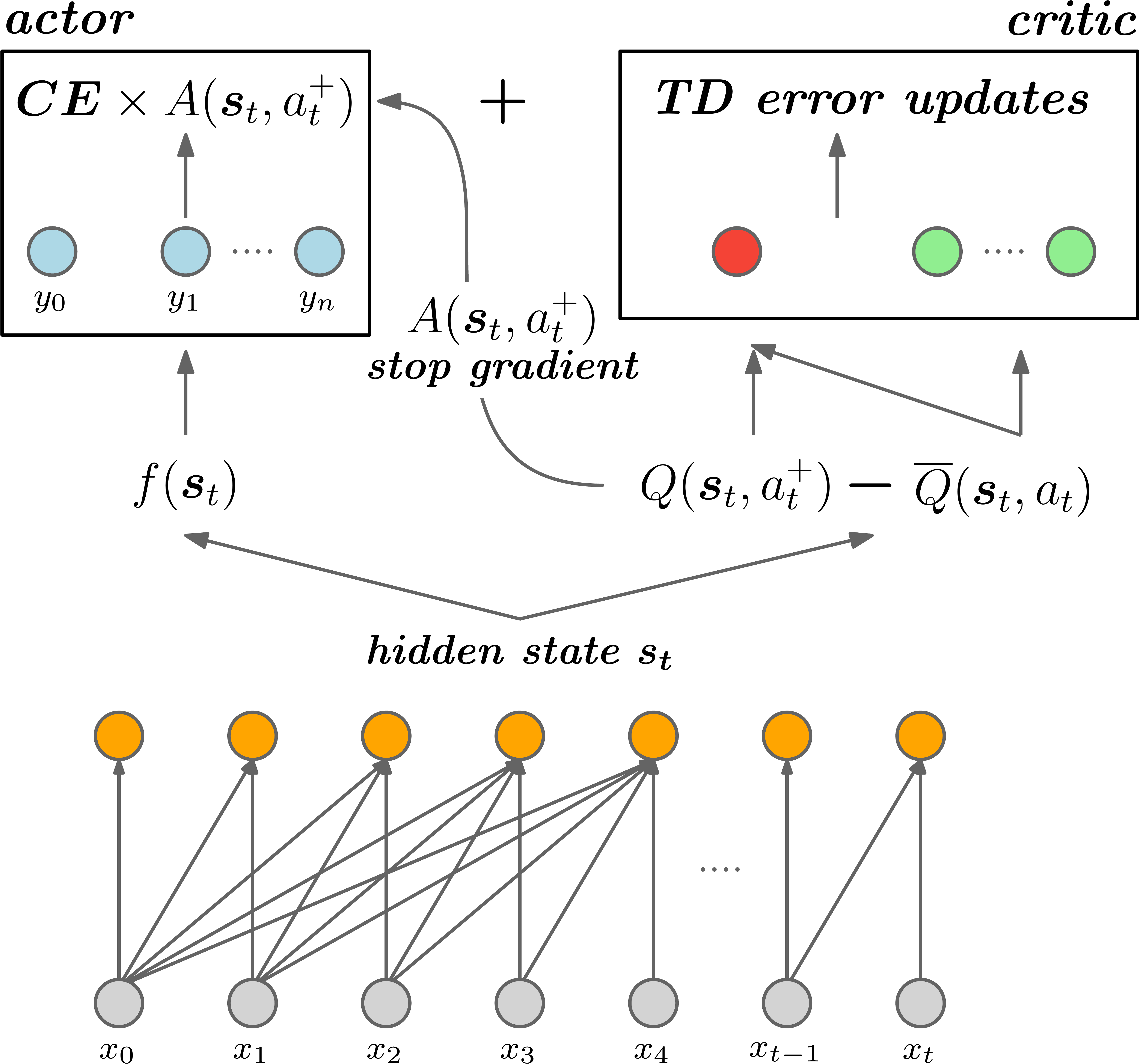}}
    \caption{The learning framework architectures of SNQN and SA2C.}
\end{figure*}

\subsection{Supervised Negative Q-learning}
Given an input user-item interaction sequence $x_{1:t}$ and an existing recommendation model $G(\cdot)$, the supervised training loss is defined as the cross-entropy over the classification distribution:
\begin{equation}
	\label{supervised-loss}
	L_s=-\sum_{i=1}^nY_ilog(p_i), \text{where } p_i=\frac{e^{y_i}}{\sum_{i'=1}^ne^{y_{i'}}}.
\end{equation}
$Y_i$ is an indicator function defined as $Y_i=1$ if the user interacted with the $i$-th item in the next timestamp, else $Y_i=0$. The cross-entropy loss pushes positive logits to high values. 
This loss provides negative learning signals by pushing down the output values of items that the user has not interacted with. This is particularly helpful in a RS setting where ranking items which are likely to be interacted by the user in the top-$k$ positions is the main goal.

Since $G(\cdot)$ already encodes the input sequence into a latent state $\mathbf{s}_t$, we can directly reuse it as the state of the RL model. This sharing schema of the base model enables the transfer of knowledge between supervised learning and RL. On the shared base model $G(\cdot)$, we formulate another output layer to map the state into Q-values:
\begin{equation}
	\label{q-value-calculation}
	Q(\mathbf{s}_t,a_t)=\delta(\mathbf{s}_t\mathbf{h}_t^T+b)=\delta(G(x_{1:t})\mathbf{h}_t^T+b),
\end{equation}
where $\delta$ denotes the activation function, $\mathbf{h}_t$ and $b$ are trainable parameters of the Q-learning output layer.
When learning from logged implicit feedback data, typically there are no negative reward signals \cite{rendle2009bpr,wrmf}. Q-learning solely based on positive reward signals (clicks, views, etc.), without negative interaction signals, leads to a model with a positive bias. Such Q-values based on only observed (positive) actions cannot be 
used for generating recommendation. To address this issue, we propose a negative reward sampling strategy for the RL training procedure. More precisely, the Q-learning loss function of SNQN is defined not only on positive action rewards but also on the sampled negative ones. We define the one-step time difference (TD) Q-loss of SNQN (Figure \ref{fig:snqn-architecture}) as:
\begin{equation}
	\label{rl-loss}
	\begin{split}
	L_q=\underbrace{(r(\mathbf{s}_t,a_t^+)+\gamma\max_{a'}Q(\mathbf{s}_{t+1},a')-Q(\mathbf{s}_t,a_t^+))^2}_{L_p:\text{ positive TD error}}\\
     +\underbrace{\sum_{a_t^-\in N_t}(r(\mathbf{s}_t,a_t^-)+\gamma\max_{a'}Q(\mathbf{s}_{t},a')-Q(\mathbf{s}_t,a_t^-))^2}_{L_n:\text{ negative TD error}},
	\end{split}
\end{equation}
where $a_t^+$ and $a_t^-$ are the positive action and negative action at timestamp $t$, respectively. $N_t$ denotes the set of sampled unobserved (negative) actions. For the negative TD error, the maximum operation is performed in $Q(\mathbf{s}_t,a')$ other than $Q(\mathbf{s}_{t+1},a')$ since we assume that taking negative actions will not affect the user state as discussed in section \ref{RL}. We assign a constant reward value $r_n$ for negative actions (i.e.,$r(\mathbf{s}_t,a_t^-)=r_n$), while the positive reward $r(\mathbf{s}_t,a_t^+)$ we can define it according to the specific demands of the recommendation domain, e.g. in e-commerce we can assign a higher reward to actions which lead to purchases rather than just clicks.
We jointly train the supervised and RL loss on the replay buffer generated from the logged implicit feedback data:
\begin{equation}
	\label{L-SQN}
	L_{snqn}=L_s+L_q.
\end{equation}
\begin{algorithm}[htp]
 \label{alg:SQN}
 \caption{Training procedure of SNQN}
 	\begin{algorithmic}[1]
        \renewcommand{\algorithmicrequire}{\textbf{Input:}} 
        \renewcommand{\algorithmicensure}{\textbf{Output:}}
        \Require
        user-item interaction sequence set $\mathcal{X}$, recommendation model $G(\cdot)$, reinforcement head $Q(\cdot)$, supervised head $f(\cdot)$, pre-defined reward function $r(\mathbf{s},a)$
        \Ensure
        all parameters in the learning space $\Theta$
        \State Initialize all trainable parameters
        \State Create $G'(\cdot)$ and $Q'(\cdot)$ as copies of $G(\cdot)$ and $Q(\cdot)$, respectively
        \Repeat 
            \State Draw a mini-batch of $(x_{1:t},a_t^+)$ from $\mathcal{X}$
            \State Draw negative actions set $N_t$ for $x_{1:t}$
            \State $\mathbf{s}_t=G(x_{1:t})$, $\mathbf{s}'_t=G'(x_{1:t})$
            \State $\mathbf{s}_{t+1}=G(x_{1:t+1})$, $\mathbf{s}'_{t+1}=G'(x_{1:t+1})$
            \State Generate random variable $z \in (0,1)$ uniformly
            \If {$z\leq0.5$}
                \State $a_*^+=\text{argmax}_a\text{ }Q(\mathbf{s}_{t+1},a)$, $a_*^-=\text{argmax}_a\text{ }Q(\mathbf{s}_{t},a)$
                \State $L_p=(r(\mathbf{s}_t,a_t^+)+\gamma Q'(\mathbf{s}'_{t+1},a_*^+)-Q(\mathbf{s}_t,a_t^+))^2$
                \State  $L_n=\sum_{a_t^-\in N_t}(r(\mathbf{s}_t,a_t^-)+\gamma Q'(\mathbf{s}'_{t},a_*^-)-Q(\mathbf{s}_t,a_t^-))^2$
                \State Calculate $L_s$ and $L_{snqn}=L_s+L_p+L_n$
                \State Perform updates by $\nabla_\Theta L_{snqn}$
            \Else
                \State $a_*^+=\text{argmax}_a\text{ }Q'(\mathbf{s}_{t+1},a)$, $a_*^-=\text{argmax}_a\text{ }Q'(\mathbf{s}_{t},a)$
                \State $L_p=(r(\mathbf{s}_t,a_t^+)+\gamma Q(\mathbf{s}_{t+1},a_*^+)-Q'(\mathbf{s}'_t,a_t^+))^2$
                \State  $L_n=\sum_{a_t^-\in N_t}(r(\mathbf{s}_t,a_t^-)+\gamma Q(\mathbf{s}_{t},a_*^-)-Q'(\mathbf{s}'_t,a_t^-))^2$
                \State Calculate $L_s$ and $L_{snqn}=L_s+L_p+L_n$
                \State Perform updates by $\nabla_\Theta L_{snqn}$
            \EndIf
        \Until converge
        \State return all parameters in $\Theta$
 	\end{algorithmic}
 \end{algorithm}
We use double Q-learning for better stability \cite{double-q-learning}, training two copies of model parameters. Algorithm 1 describes the training procedure. Figure \ref{fig:snqn-architecture} shows the SNQN architecture.
\subsection{Supervised Advantage Actor-Critic}
Actor-Critic (AC) methods have been successfully used in RL. The key idea of AC methods is the introduction of a critic that evaluates the goodness of an action taken and assigns higher weights to actions with high cumulative rewards. In the SNQN method, the supervised component can be seen as the actor which aims at imitating the logged user behavior. A simple solution for the critic is to use the output Q-values from the RL head, as these Q-values measure the cumulative rewards the system gains given the state-action pair. These Q-values are sensitive to the specific timestamp of the sequence, a bad action in an early timestamp of a long sequence could also have a high Q-value since Q-values are based on the cumulative gains of all the following actions in this sequence.

Instead of the absolute Q-value, what we actually would like to measure is how much ``advantage'' we obtain by applying an action, compared to the average case (i.e. average Q-values). This advantage can help us alleviate the bias introduced from the sequence timestamp. However, calculating the average Q-values along the whole action space would introduce additional computation cost, especially when the candidate item set is large. To this end, we have introduced negative samples in the SNQN method. A concise solution is to calculate the average among the sampled actions (including both positive and negative examples) as an approximation. Based on this motivation, the average Q-values can be defined as:
\begin{equation}
	\label{eq:average} \overline{Q}(\mathbf{s}_t,a)=\frac{\sum_{a'\in a_t^+\cap N_t}Q(\mathbf{s}_t,a')}{|N_t|+1}.
\end{equation}
The advantage of an observed (positive) action is formulated as:
\begin{equation}
	\label{eq:advantage} A(\mathbf{s}_t,a_t^+)=Q(\mathbf{s}_t,a_t^+)-\overline{Q}(\mathbf{s}_t,a).
\end{equation}
We use this advantage to re-weight the actor (i.e. the supervised head). If a positive action has higher advantage than the average, we increase its weight, and vice versa. To enhance stability, we stop the gradient flow and fix the Q-values when they are used to calculate the average and advantage. We then train the actor and critic jointly. The training loss of SA2C is formulated as: 
\begin{equation}
	\label{eq:L_SA2C}
	L_{sa2c}=L_a+L_q,\text{ where }L_a=L_s\cdot A(\mathbf{s}_t,a_t^+).
\end{equation}
Figure \ref{fig:sa2c-architecture} illustrates the architecture of SA2C. During the training procedure, the learning of Q-values can be unstable \cite{parisotto2019stabilizing}, particularly in the early stage. To mitigate these issues, we pre-train the model using SNQN in the first $T$ training steps (batches).
When the Q-values become more stable, we start to use the advantage to re-weight the actor and perform updates according to the architecture of Figure \ref{fig:sa2c-architecture}. We  use double Q-learning and the training procedure of SA2C is similar to Algorithm 1 except for the computation of advantage and the re-weighting of $L_s$.
\section{Experiments}\label{sec:exp}
We conduct experiments \footnote{The implementation code and data can be found at \url{https://drive.google.com/file/d/185KB520pBLgwmiuEe7JO78kUwUL_F45t/view?usp=sharing}} 
on two real-world datasets to evaluate SNQN and SA2C in the e-commerce scenario. Both datasets contain click and purchase interactions. We use the supervised head to generate recommendations without special mention. We address the following research questions:

\textbf{RQ1:} How do the proposed methods perform when integrated with different base models?
	
\textbf{RQ2} What is the performance if we use the Q-leaning head to generate recommendation?

\textbf{RQ3:} What is the performance if we introduce an additional off-policy correction term in the actor of SA2C?

\textbf{RQ4:} How does the negative sampling strategy affect the performance?

\subsection{Experimental Settings}
\subsubsection{Datasets:}
 RC15\footnote{\url{https://recsys.acm.org/recsys15/challenge/}} and RetailRocket\footnote{\url{https://www.kaggle.com/retailrocket/ecommerce-dataset}}. 

\textbf{RC15.} This is based on the dataset of RecSys Challange 2015. The dataset is session-based and each session contains a sequence of clicks and purchases. We remove sessions whose length is smaller than 3 and then sample a subset of 200k sessions.

\textbf{RetailRocket.} This dataset is collected from a real-world e-commerce website. It contains session events of viewing and adding to cart. To keep in line with the RC15 dataset, we treat views as clicks and adding to cart as purchases. We remove the items which are interacted less than 3 times and the sequences whose length is smaller than 3. 

Table \ref{Datasets} summarizes the statistics of the two datasets.
\begin{table}
    \centering
    \setlength{\abovecaptionskip}{3pt}
    \begin{threeparttable}
    \caption{Dataset statistics.}
    \label{Datasets}
    \begin{tabular}{p{1.5cm}p{2.0cm}<{\centering}p{2.0cm}<{\centering}}
    \toprule
    Dataset&RC15 & RetailRocket\cr
    \midrule
    \#sequences&200,000&195,523\cr
    \#items&26,702&70,852\cr
    \#clicks&1,110,965&1,176,680\cr
    \#purchase&43,946&57,269\cr
    \bottomrule
  \end{tabular}
    \end{threeparttable}
\end{table}
\subsubsection{Evaluation protocols}
We adopt cross-validation to evaluate the performance of the proposed methods. The ratio of training, validation, and test set is 8:1:1. We randomly sample 80\% of sequences as the training set. Each experiment is repeated five times, and the average performance is reported.

The recommendation quality is measured with two metrics: Hit Ratio (HR) and Normalized Discounted Cumulative Gain (NDCG). HR@$k$ is a recall-based metric, measuring whether the ground-truth item is in the top-$k$ positions of the recommendation list. We can define HR for clicks as:
\begin{equation}
	\label{HR_click}
	\text{HR(click)}=\frac{\#\text{hits among clicks}}{\#\text{clicks in test}}.
\end{equation}
HR(purchase) is defined similarly with HR(click) by replacing the clicks with purchases. NDCG is a rank sensitive metric which assign higher scores to top positions in the recommendation list \cite{NDCG}.

As we focus on the e-commerce scenario, we assign a higher reward to actions leading to purchases (i.e. conversions) compared to actions leading to only clicks. 
If a recommended item is not interacted with by the user, we give this action a zero reward. 
Hence the cumulative reward for evaluation is proportional to HR.

\subsubsection{Baselines}
We integrated the proposed SNQN and SA2C with four state-of-the-art sequential recommendation models:
\begin{itemize}
    \item GRU \cite{gru4rec}: This method utilizes a GRU to model the input sequences. The final hidden state of the GRU is treated as the latent representation for the input sequence.
	\item Caser \cite{caser-rec}: This is a recently proposed CNN-based method, which captures sequential signals by applying convolution operations on the embedding matrix of previous items.
	\item NItNet \cite{nextitnet}: NItNet uses dilated CNN for larger receptive field and residual connection to increase network depth.
	\item SASRec \cite{SASRec}: This baseline is based on self-attention and uses the Transformer \cite{Transformer} architecture. The output of the Transformer encoder is treated as the latent sequence state. 
\end{itemize}
We compare SNQN, SA2C with SQN, SAC\cite{xin2020self}, respectively.

\subsubsection{Parameter settings}
For both datasets, the input sequences are composed of 10 interacted items. If the sequence length is less than 10, we complement the sequence with a padding item.
We train all models with the Adam optimizer \cite{kingma2014adam}. The mini-batch size is 256. For SNQN, the learning rate is 0.01 on RC15 and 0.005 on RetailRocket, which is the identical to SQN \cite{xin2020self}. For SA2C, we use the same learning rate with SNQN at the early pre-training stage. After that, the learning rate is set as 0.001 on both datasets. 
We use the basic uniform distribution for negative sampling strategy to eliminate any influence from the sampler.
The item embedding size is set to 64 for all models. For GRU, the size of the hidden state is 64. For Caser, we use 1 vertical convolution filter and 16 horizontal filters whose heights are set from \{2,3,4\}. The drop-out ratio is set to 0.1. For NextItNet, we use the published implementation \cite{nextitnet} with the predefined settings. For SASRec, the number of heads in self-attention is set as 1, according to the original paper \cite{SASRec}.  When SNQN and SA2C are integrated with a base model, the hyper-parameter setting of the base model remains exactly unchanged.

For the training of SNQN and SA2C, the discount factor $\gamma$ is set as 0.5. The ratio between the click reward ($r_c$) and the purchase reward ($r_p$) is set as $r_p/r_c=5$. These settings are the same as in \cite{xin2020self} for a fair comparison.
If without special mention, for one positive action we sample 10 negative actions in the training procedure.  The reward for negative actions is set as $r_n=0$.

\subsection{Performance Comparison (RQ1)}
\begin{table*}
    \centering
    \setlength{\abovecaptionskip}{3pt}
    \begin{threeparttable}
    \caption{Top-$k$ recommendation performance comparison of different models ($k=5, 10, 20$) on RC15 dataset. Recommendations are generated from the supervised head. NG is short for NDCG. Boldface denotes the highest score.
    }
    \label{comparison between different models on RC15}
    \begin{tabular}{p{2.0cm}p{0.85cm}<{\centering}p{0.85cm}<{\centering}p{0.85cm}<{\centering}p{0.85cm}<{\centering}p{0.85cm}<{\centering}p{0.85cm}<{\centering}p{0.85cm}<{\centering}p{0.85cm}<{\centering}p{0.85cm}<{\centering}p{0.85cm}<{\centering}p{0.85cm}<{\centering}p{0.85cm}<{\centering}}
    \toprule
    \multirow{2}{*}{Models}&\multicolumn{6}{c}{purchase}&\multicolumn{6}{c}{click}\cr
    \cmidrule(lr){2-7} \cmidrule(lr){8-13}
    &HR@5&NG@5&HR@10&NG@10&HR@20&NG@20&HR@5&NG@5&HR@10&NG@10&HR@20&NG@20\cr
    \midrule
    GRU    &0.3994&0.2824 &0.5183&0.3204 &0.6067&0.3429 &0.2876&0.1982 &0.3793&0.2279 &0.4581&0.2478\cr
    \hdashline
    GRU-SQN&$0.4228$&$0.3016$&$0.5333$&$0.3376$&$0.6233$&$0.3605$&$0.3020$ &$0.2093$ &$0.3946$&$0.2394$&0.4741&0.2587\cr
    GRU-SNQN&0.4368&0.3115&0.5428&0.3460&0.6316&0.3686&0.3124&0.2164&0.4067&0.2469&0.4856&0.2669\cr
    \hdashline
    GRU-SAC&$0.4394$&$0.3154$&0.5525&0.3521&$0.6378$&$0.3739$&0.2863 &0.1985 &0.3764 &0.2277 &0.4541&0.2474\\ 
    GRU-SA2C&\textbf{0.4514}&\textbf{0.3297}&\textbf{0.5606}&\textbf{0.3652}&\textbf{0.6420}&\textbf{0.3859}&\textbf{0.3287}&\textbf{0.2307}&\textbf{0.4214}&\textbf{0.2606}&\textbf{0.5000}&\textbf{0.2806}\cr
   \hline
    Caser& 0.4475&0.3211 &0.5559&0.3565&0.6393&0.3775&0.2728 &0.1896&0.3593&0.2177&0.4371&0.2372\cr
    \hdashline
    Caser-SQN&$0.4553$&$0.3302$&$0.5637$&$0.3653$&$0.6417$&$0.3862$&0.2742&0.1909 &0.3613&0.2192&0.4381&0.2386\cr
    Caser-SNQN&0.4781&0.3460&0.5876&0.3816&0.6657&0.4015&0.2800&0.1951&0.3682&0.2237&0.4465&0.2436\cr
    \hdashline
    Caser-SAC&$0.4866$&$0.3527$&$0.5914$&0.3868&$0.6689$&$0.4065$&0.2726&0.1894&0.3580&0.2171&0.4340&0.2362\\ 
    Caser-SA2C&\textbf{0.4917}&\textbf{0.3635}&\textbf{0.6000}&\textbf{0.3989}&\textbf{0.6796}&\textbf{0.4192}&\textbf{0.2948}&\textbf{0.2068}&\textbf{0.3835}&\textbf{0.2356}&\textbf{0.4596}&\textbf{0.2549}\cr
    \hline
    NItNet&0.3632&0.2547&0.4716&0.2900&0.5558&0.3114&0.2950&0.2030&0.3885&0.2332&0.4684&0.2535\cr
    \hdashline
    NItNet-SQN&$0.3845$&$0.2736$&$0.4945$&$0.3094$&$0.5766$&$0.3302$&$0.3091$ &$0.2137$ &$0.4037$ &$0.2442$&$0.4835$&0.2645\cr
    NItNet-SNQN&0.3969&0.2803&0.5039&0.3152&0.5876&0.3363&0.3153&0.2176&0.4098&0.2482&0.4896&0.2686\cr
    \hdashline
    NItNet-SAC&$0.3914$&$0.2813$&$0.4964$&$0.3155$&$0.5763$&$0.3357$&$0.2977$&$0.2055$&0.3906&$0.2357$&0.4693&$0.2557$\\ 
    NItNet-SA2C&\textbf{0.4382}&\textbf{0.3171}&\textbf{0.5403}&\textbf{0.3505}&\textbf{0.6259}&\textbf{0.3722}&\textbf{0.3410}&\textbf{0.2395}&\textbf{0.4348}&\textbf{0.2699}&\textbf{0.5113}&\textbf{0.2897}\cr
    \hline
    SASRec& 0.4228&0.2938 &0.5418&0.3326&0.6329&0.3558&0.3187&0.2200&0.4164&0.2515&0.4974&0.2720\cr
    \hdashline
    SASRec-SQN&0.4336&$0.3067$&0.5505&$0.3435$& $0.6442$&$0.3674$ &$0.3272$&$0.2263$&$0.4255$&$0.2580$&$0.5066$&$0.2786$\cr
    SASRec-SNQN&0.4435&0.3163&0.5581&0.3535&0.6450&0.3742&0.3284&0.2267&0.4271&0.2588&0.5083&0.2794\cr
    \hdashline
    SASRec-SAC&$0.4540$&$0.3246$&$0.5701$&$0.3623$&$0.6576$& $0.3846$&0.3130&0.2161&0.4114&0.2480&0.4945&0.2691\cr
    SASRec-SA2C&\textbf{0.4705}&\textbf{0.3385}&\textbf{0.5756}&\textbf{0.3728}&\textbf{0.6648}&\textbf{0.3956}&\textbf{0.3444}&\textbf{0.2407}&\textbf{0.4402}&\textbf{0.2719}&\textbf{0.5194}&\textbf{0.2920}\\
    \bottomrule
    \end{tabular}
    \end{threeparttable}
    \vspace{-0.1cm}
\end{table*}

\begin{figure}
    \captionsetup[subfloat]
    {}
    \centering
    \subfloat[Purchase predictions]{
    \label{fig:conv-purchase-rc}
    \includegraphics[width=0.23\textwidth]{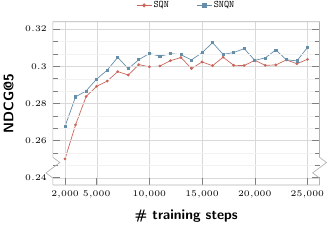}}
    \hspace{0.5mm}
    \subfloat[Click predictions]{%
    \label{fig:conv-click-rc}
    \includegraphics[width=0.23\textwidth]{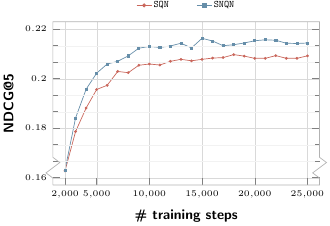}}
    \vspace{-0.2cm}
    \caption{Model convergence comparison on RC15}
    \label{fig:model-conv}
    \vspace{-0.2cm}
\end{figure}

Table \ref{comparison between different models on RC15} and Table \ref{comparison between different models on RetailRocket} show the performance of top-$k$ recommendations on RC15 and RetailRocket, respectively. 

(1) The introduced negative sampling strategy on the RL head does improve the learning performance also on the supervised component. This can be attributed to the shared recommendation model $G(\cdot)$ between the supervised part and the RL part. We also observe that SNQN achieves faster convergence than SQN. Figure \ref{fig:model-conv} shows the comparison between model convergence under the same learning rate on the validation set of RC15, using GRU as the base model $G(\cdot)$. Results on RetailRocket and other base models lead to the same conclusion. This further demonstrates that negative sampling helps the model to learn faster and improves its performance.

(2) SA2C achieves better performance than SAC in most cases. This indicates that the advantage estimate used in SA2C is a more effective critic compared with the raw Q-values used in SAC. This can be attributed to the fact that the advantage estimation helps to alleviate the sequence timestamp bias.

(3) SA2C always achieves the highest NDCG. This is due to the fact that positive actions are weighted (advantaged) in a more effective manner during the training procedure of SA2C.

\begin{table*}
    \centering
    \setlength{\abovecaptionskip}{3pt}
    \begin{threeparttable}
    \caption{Top-$k$ recommendation performance comparison of different models ($k=5, 10, 20$) on RetailRocket. Recommendations are generated from the supervised head. NG is short for NDCG. Boldface denotes the highest score. 
    }
    \label{comparison between different models on RetailRocket}
    \begin{tabular}{p{2.0cm}p{0.85cm}<{\centering}p{0.85cm}<{\centering}p{0.85cm}<{\centering}p{0.85cm}<{\centering}p{0.85cm}<{\centering}p{0.85cm}<{\centering}p{0.85cm}<{\centering}p{0.85cm}<{\centering}p{0.85cm}<{\centering}p{0.85cm}<{\centering}p{0.85cm}<{\centering}p{0.85cm}<{\centering}}
    \toprule
    \multirow{2}{*}{Models}&\multicolumn{6}{c}{purchase}&\multicolumn{6}{c}{click}\cr
    \cmidrule(lr){2-7} \cmidrule(lr){8-13}
    &HR@5&NG@5&HR@10&NG@10&HR@20&NG@20&HR@5&NG@5&HR@10&NG@10&HR@20&NG@20\cr
    \midrule
    GRU    &0.4608 &0.3834&0.5107 &0.3995&0.5564&0.4111&0.2233&0.1735&0.2673&0.1878&0.3082&0.1981\cr
    \hdashline
    GRU-SQN&$0.5069$&$0.4130$&$0.5589$&$0.4289$&$0.5946$&$0.4392$&$0.2487$ &$0.1939$ &0.2967&$0.2094$&$0.3406$&$0.2205$\cr
    GRU-SNQN&0.5232&0.4376&0.5713&0.4544&0.6175&0.4650&0.2662&0.2065&0.3181&0.2233&\textbf{0.3656}&0.2353\cr
    \hdashline
    GRU-SAC&$0.4942$&$0.4179$&$0.5464$&$0.4341$&$0.5870$&$0.4428$&$0.2451$&$0.1924$&$0.2930$ &$0.2074$&$0.3371$&$0.2186$\\
    GRU-SA2C&\textbf{0.5526}&\textbf{0.4754}&\textbf{0.5963}&\textbf{0.4897}&\textbf{0.6313}&\textbf{0.4985}&\textbf{0.2720}&\textbf{0.2150}&\textbf{0.3208}&\textbf{0.2308}&\textbf{0.3656}&\textbf{0.2422}\cr
    \hline
    Caser& 0.3491&0.2935 &0.3857&0.3053&0.4198&0.3141&0.1966&0.1566&0.2302&0.1675&0.2628&0.1758\cr
    \hdashline
    Caser-SQN&$0.3674$&$0.3089$&$0.4050$&$0.3210$&$0.4409$&$0.3301$&$0.2089$&$0.1661$ &$0.2454$&$0.1778$&$0.2803$&$0.1867$\cr
    Caser-SNQN&0.3757&0.3179&0.4181&0.3317&0.4595&0.3422&0.2160&0.1721&0.2530&0.1841&0.2895&0.1934\cr
    \hdashline
    Caser-SAC&$0.3871$&$0.3234$&$0.4336$&$0.3386$&$\mathbf{0.4763}$&$0.3494$&$\textbf{0.2206}$&$0.1732$&$\textbf{0.2617}$&$0.1865$&$\textbf{0.2999}$&$0.1961$\\ 
    Caser-SA2C&\textbf{0.3971}&\textbf{0.3446}&\textbf{0.4381}&\textbf{0.3578}&0.4733&\textbf{0.3667}&0.2170&\textbf{0.1759}&0.2528&\textbf{0.1875}&0.2873&\textbf{0.1963}\cr
    \hline
    NItNet&0.5630&0.4630&0.6127&0.4792&0.6477&0.4881&0.2495&0.1906&0.2990&0.2067&0.3419&0.2175\cr
    \hdashline
    NItNet-SQN&$0.5895$&$0.4860$&$0.6403$&$0.5026$&$0.6766$&$0.5118$&$0.2610$ &0.1982 &0.3129&$0.2150$&$0.3586$&$0.2266$\cr
    NItNet-SNQN&0.6016&0.5062&0.6543&0.5234&\textbf{0.6921}&0.5330&0.2699&0.2065&0.3236&0.2240&0.3703&0.2358\cr
    \hdashline
    NItNet-SAC&$0.5895$&$0.4985$&$0.6358$&$0.5162$&$0.6657$&$0.5243$&$0.2529$&$0.1964$&$0.3010$&$0.2119$&$0.3458$&$0.2233$\\ 
    NItNet-SA2C&\textbf{0.6226}&\textbf{0.5422}&\textbf{0.6573}&\textbf{0.5534}&0.6842&\textbf{0.5603}&\textbf{0.2787}&\textbf{0.2197}&\textbf{0.3271}&\textbf{0.2354}&\textbf{0.3719}&\textbf{0.2468}\cr
    \hline
    SASRec&0.5267&0.4298&0.5916&0.4510&0.6341&0.4618&0.2541&0.1931&0.3085&0.2107&0.3570&0.2230\cr
    \hdashline
    SASRec-SQN&$0.5681$&$0.4617$&$0.6203$&$0.4806$&$0.6619$&$0.4914$&$0.2761$&$0.2104$&$0.3302$&$0.2279$&$0.3803$&$0.2406$\cr
    SASRec-SNQN&0.5776&0.4846&0.6310&0.5020&0.6719&0.5123&0.2815&0.2171&0.3381&0.2355&0.3888&0.2483\cr
    \hdashline
    SASRec-SAC&$0.5623$&$0.4679$&$0.6127$&$0.4844$&$0.6505$& $0.4940$&$0.2670$&$0.2056$&$0.3208$&$0.2230$&$0.3701$&$0.2355$\cr
    SASRec-SA2C&\textbf{0.5929}&\textbf{0.5080}&\textbf{0.6437}&\textbf{0.5246}&\textbf{0.6798}&\textbf{0.5337}&\textbf{0.2873}&\textbf{0.2242}&\textbf{0.3409}&\textbf{0.2416}&\textbf{0.3893}&\textbf{0.2538}
    \\
    \bottomrule
    \end{tabular}
    \end{threeparttable}
\end{table*}
\subsection{Recommendations from Q-learning (RQ2)}
Table \ref{recommendation from Q} shows the performance comparison when we use the Q-learning head to generate recommendations. We compare the performance of SNQN with a simple double Q-learning (DQN) algorithm with the same negative sampling strategy but without a supervised head upon the base model. The performance of SA2C is not significantly different with SNQN as the two methods are essentially identical with regards to the Q-learning head. We use the same base model GRU and the same hyper-parameters for DQN and SNQN. Results on the other base models show identical trends. We observe that SNQN achieves better performance than DQN in all evaluation metrics on both purchase and click predictions. Combined with the results of Table \ref{comparison between different models on RC15} and Table \ref{comparison between different models on RetailRocket}, we observe that joint training of supervised learning and RL with shared base models helps to improve the performance of each component. 

\subsection{Effect of Off-Policy Correction (RQ3)}
\begin{table}
    \centering
    \setlength{\abovecaptionskip}{3pt}
    \begin{threeparttable}
    \caption{Recommendation from the RL head. Boldface denotes the highest score. DQN denotes only a Q-learning head is used without the supervised head. 
    }
    \label{recommendation from Q}
    \begin{tabular}{p{0.8cm}p{1.5cm}<{\centering}p{0.85cm}<{\centering}p{0.85cm}<{\centering}p{0.85cm}<{\centering}p{0.85cm}<{\centering}}
    \toprule
    &\multirow{2}{*}{Methods}&\multicolumn{2}{c}{purchase}&\multicolumn{2}{c}{click}\cr
    \cmidrule(lr){3-4} \cmidrule(lr){5-6}
    &&HR@5&NG@5&HR@5&NG@5\cr
    \midrule
    \multirow{2}{*}{RC15}&DQN&0.3642&0.2476&0.2096&0.1353\cr
    &SNQN&\textbf{0.3698}&\textbf{0.2497}&\textbf{0.2286}&\textbf{0.1495}\cr
    \midrule
    \multirow{2}{*}{\parbox{1cm}{Retail Rocket}}&DQN&0.2952	&0.2204&0.1368&0.0961\cr
    &SNQN&\textbf{0.3124}&\textbf{0.2422}&\textbf{0.1546}&\textbf{0.1103}\cr
    \bottomrule
    \end{tabular}
    \end{threeparttable}
\end{table}
\begin{table}
    \centering
    \setlength{\abovecaptionskip}{3pt}
    \begin{threeparttable}
    \caption{Effect of off-policy correction. w/o means without off-policy correction in the actor while w means the opposite. Boldface denotes the highest score.
    }
    \label{effect of off-policy correction}
    \begin{tabular}{p{0.8cm}p{1.2cm}<{\centering}p{0.9cm}<{\centering}p{0.9cm}<{\centering}p{0.9cm}<{\centering}p{0.9cm}<{\centering}}
    \toprule
    &\multirow{2}{*}{Methods}&\multicolumn{2}{c}{purchase}&\multicolumn{2}{c}{click}\cr
    \cmidrule(lr){3-4} \cmidrule(lr){5-6}
    &&NDCG&$NG_{off}$&NDCG&$NG_{off}$\cr
    \midrule
    \multirow{2}{*}{RC15}&w/o&\textbf{0.3652}&\textbf{0.1077}&\textbf{0.2606}&0.0767\cr
    &w&0.3551&0.1064&0.2595&\textbf{0.0781}\cr
    \midrule
    \multirow{2}{*}{\parbox{1cm}{Retail Rocket}}&w/o&\textbf{0.4897}&\textbf{0.2171}&\textbf{0.2308}&0.0861\cr
    &w&0.4771&0.2147&0.2238&\textbf{0.0872}\cr
    \bottomrule
    \end{tabular}
    \end{threeparttable}
\end{table}
\citet{googlewsdmoffpolicycorrection} introduced an off-policy correction term (propensity score) for the policy-gradient method. The propensity score is defined as $\rho=\frac{\pi_\theta(a|\mathbf{s})}{\beta(a|\mathbf{s})}$. In this subsection, we investigate the effect of this propensity score when introduced into the actor component of SA2C. In that case, the training loss of the actor becomes:
\begin{equation}
	\label{eq:L_a-off}
	L_{a-off}=L_s\cdot A(\mathbf{s}_t,a_t^+) \cdot \rho.
\end{equation}
We also introduce another NDCG-based off-policy corrected evaluation metric \cite{Vlassis2019off-polciy-eval} which is formulated as
\begin{equation}
	\label{eq:off-policy eval}
	NG_{off}=\frac{\sum\frac{NDCG}{\beta}}{\sum\frac{1}{\beta}}.
\end{equation}
In this implementation, we use the item frequency to approximate the behavior policy $\beta$, which is also adopted in \cite{pop_behavior}.
Table \ref{effect of off-policy correction} shows the result when generating top-10 recommendations with GRU as the base model. Results on the other base models lead to the same conclusion. We note the following observations:

\begin{figure*}
    \captionsetup[subfloat]
    {}
    \centering
    \subfloat[SNQN for purchase]{
    \label{snqn-purchase-rc-negative}
    \includegraphics[width=0.24\textwidth]{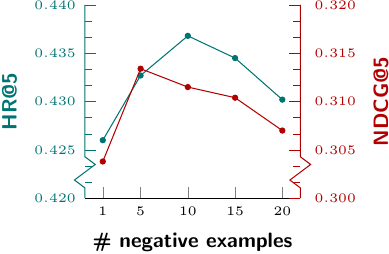}}
    \hspace{0.5mm}
    \subfloat[SNQN for click]{%
    \label{snqn-click-rc-negative}
    \includegraphics[width=0.24\textwidth]{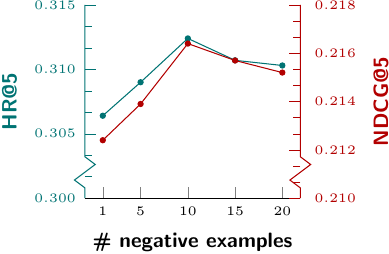}}
    \hspace{0.5mm}
    \subfloat[SA2C for purchase]{%
    \label{sa2c-purchase-rc-negative}
    \includegraphics[width=0.24\textwidth]{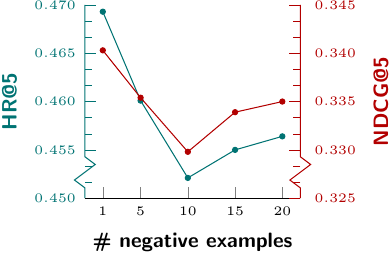}}
    \hspace{0.5mm}
    \subfloat[SA2C for click]{%
    \label{sa2c-click-rc-negative}
    \includegraphics[width=0.24\textwidth]{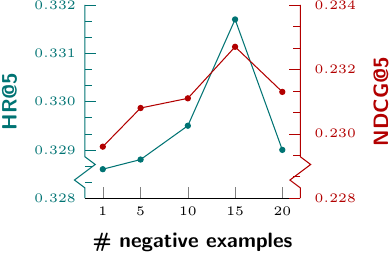}}
    \caption{Effect of number of negative samples on RC15}
    \label{negative-sample-RC}
\end{figure*}

\begin{figure*}
    \captionsetup[subfloat]
    {}
    \centering
    \subfloat[SNQN for purchase]{
    \label{snqn-purchase-retail-negative}
    \includegraphics[width=0.24\textwidth]{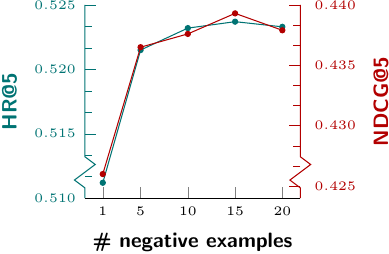}}
    \hspace{0.5mm}
    \subfloat[SNQN for click]{%
    \label{snqn-click-retail-negative}
    \includegraphics[width=0.24\textwidth]{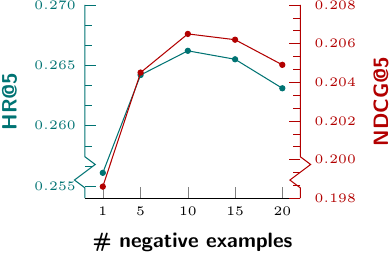}}
    \hspace{0.5mm}
    \subfloat[SA2C for purchase]{%
    \label{sa2c-purchase-retail-negative}
    \includegraphics[width=0.24\textwidth]{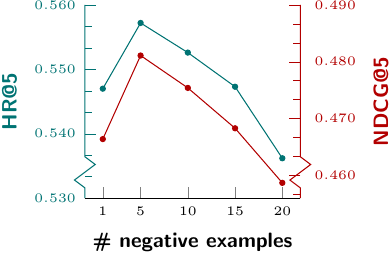}}
    \hspace{0.5mm}
    \subfloat[SA2C for click]{%
    \label{sa2c-click-retail-negative}
    \includegraphics[width=0.24\textwidth]{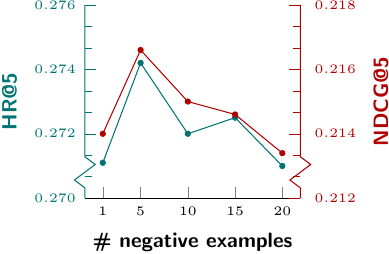}}
    \caption{Effect of number of negative samples on RetailRocket}
    \label{negative-sample-Kaggle}
\end{figure*}
(1) Off-policy correction doesn't improve the (standard) NDCG score. NDCG is actually defined on non-corrected data, so the non-corrected actor performs better at this evaluation metric.

(2) On $NG_{off}$, the off-policy correction helps the model to achieve better performance for click predictions but not for purchases. The reason for this is that clicks account for the biggest part of the dataset. Hence the off-policy correction term is actually better defined to correct the click data, leading to a better performance of $NG_{off}$ for clicks, while the high variance of the off-policy correction for the small portion of purchase data leads to less of an improvement. 
This observation indicates that perhaps we should design different corrections for different kinds of interactions.

We found that computing the off-policy correction term involves a lot of normalization techniques (e.g., clipping and smoothing) \cite{googlewsdmoffpolicycorrection}. The behavior policy $\beta$ can also be a long-tail distribution \cite{pop_behavior}. This introduces substantial noise and high variance into the training procedure. Designing more effective and stable off-policy correction terms remains an open research problem.

\subsection{Hyperparameter Study (RQ4)}
We conduct a series of experiments to demonstrate the effect of negative sampling on the RL component. Figure \ref{negative-sample-RC} and Figure \ref{negative-sample-Kaggle} show the recommendation accuracy with different sizes of negative examples (i.e. $|N_t|$) on RC15 and RetailRocket, respectively (the base model is GRU). On both click and purchase predictions, the recommendation performance initially increases and then decreases (except in Figure \ref{sa2c-purchase-rc-negative}).
When more negative actions are introduced, the model has more data to learn from. By introducing negative actions, the model does not only learn that actions leading to purchases are better than actions leading to clicks, but also learns to draw a contrast between negative (uninteracted) and positive actions. Increasing the sample size means that the model can have access to more diverse negative signals and, thus, leads to better performance. 
In Figure \ref{sa2c-purchase-rc-negative}, we observe that the model achieves a good performance with small sample sizes. A small sample could introduce more noise into the estimation of the advantage and may help the model to find a better local optimal with higher performance with more update steps to converge. We have observed in the experiments that SA2C needs more iterations to converge when the sample size is small. 

Table \ref{effect of negative rewards settings} shows the effect of different negative reward settings (i.e., $r_n$) on RC15 dataset when using GRU as the base model. Results on RetailRocket lead to the same conclusion. 
$r_n$ can be seen as the strength of negative signals. We can see from Table \ref{effect of negative rewards settings} that different $r_n$ settings make no significant difference regarding the recommendation performance. However, through the performance comparison between SNQN and SQN, we find that the presence of negative samples in the RL training procedure dramatically affects the recommendation accuracy. This fits with the finding in \cite{reddy2019sqil,xin-etal-2018-batch}.
\begin{table}
    \centering
    \setlength{\abovecaptionskip}{3pt}
    \begin{threeparttable}
    \caption{Effect of negative rewards settings on RC15.
    }
    \label{effect of negative rewards settings}
    \begin{tabular}{p{0.8cm}p{1.2cm}<{\centering}p{0.9cm}<{\centering}p{0.9cm}<{\centering}p{0.9cm}<{\centering}p{0.9cm}<{\centering}}
    \toprule
    &\multirow{2}{*}{$r_n$}&\multicolumn{2}{c}{purchase}&\multicolumn{2}{c}{click}\cr
    \cmidrule(lr){3-4} \cmidrule(lr){5-6}
    &&HR@5&NG@5&HR@5&NG@5\cr
    \midrule
    \multirow{4}{*}{SNQN}&0&0.4368&0.3115&0.3124&0.2164\cr
    &-0.5&0.4324&0.3043&0.3118&0.2158\cr
    &-1.0&0.4345&0.3091&0.3108&0.2160\cr
    &-2.0&0.4269&0.3072&0.3128&0.2173\cr
    \midrule
    \multirow{4}{*}{SA2C}&0&0.4514&0.3297&0.3287&0.2307\cr
    &-0.5&0.4479&0.3263&0.3326&0.2332\cr
    &-1.0&0.4486&0.327&0.332&0.2333\cr
    &-2.0&0.4511&0.3274&0.3321&0.2335\cr
    \bottomrule
    \end{tabular}
    \end{threeparttable}
    \vspace{-0.2cm}
\end{table}
\vspace{-0.1cm}
\section{Conclusion}
In this paper, we propose two learning frameworks (SNQN and SA2C) to explore the usage of RL under recommendation settings. SNQN combines supervised learning and RL with the shared base model and introduces negative sampling into the RL training procedure. The explicitly introduced negative comparison signals help the RL output layer to perform good ranking.
Based on the sampled actions, SA2C first computes the advantage of actions which can be seen as normalized Q-values and then use this advantage estimate as a critic to re-weight the actor.
To verify the effectiveness of our methods, we integrate them into four state-of-the-art recommendation models and conduct experiments on two real-world e-commerce datasets. Our experimental findings demonstrate that the proposed SNQN and SA2C are effective in further improving the recommendation performance, compared to existing self-supervised RL methods.
\newpage

\bibliographystyle{ACM-Reference-Format}
\bibliography{sample-bibliography}

\end{document}